\documentclass[runningheads]{llncs}
\usepackage{paper}
\usepackage{url}
\urldef{\maillars}\path|lars.patrick.hillebrand@iais.fraunhofer.de|
\urldef{\mailsdavid}\path|david.biesner@iais.fraunhofer.de|

\begin{document}
\title{Interpretable Topic Extraction and Word Embedding Learning using row-stochastic DEDICOM}
\titlerunning{Topic Extraction and Word Embedding Learning using DEDICOM}
%

\author{Lars Hillebrand*\inst{1,2} \and
David Biesner*\inst{1,2} \and
Christian Bauckhage\inst{1,2} \and
Rafet Sifa\inst{1}}
\authorrunning{Hillebrand, Biesner et al.}

%
\institute{Fraunhofer IAIS \and
University of Bonn}

\newenvironment{symbolfootnotes}
  {\par\edef\savedfootnotenumber{\number\value{footnote}}
  \renewcommand{\thefootnote}{\fnsymbol{footnote}}
  \setcounter{footnote}{0}}
  {\par\setcounter{footnote}{0}}
\begin{symbolfootnotes}
\footnotetext[1]{First authors, equal contribution. \\ 
Correspondence to \email{lars.patrick.hillebrand@iais.fraunhofer.de}}
\end{symbolfootnotes}

\maketitle              

\begin{abstract}

The DEDICOM algorithm provides a uniquely interpretable matrix factorization method for symmetric and asymmetric square matrices.
We employ a new row-stochastic variation of DEDICOM on the pointwise mutual information matrices of text corpora to identify latent topic clusters within the vocabulary and simultaneously learn interpretable word embeddings.
We introduce a method to efficiently train a constrained DEDICOM algorithm and a qualitative evaluation of its topic modeling and word embedding performance.

\end{abstract}

\section{Introduction}
Matrix factorization methods have always been a staple in many natural language processing (NLP) tasks. Factorizing a matrix of word co-occurrences can create both low-dimensional representations of the vocabulary, so-called word embeddings \cite{word2vecpmi,glove}, that carry semantic and topical meaning within them, as well as representations of meaning that go beyond single words to latent topics.

\begin{figure}
    \centering
    \includegraphics[width=0.5\textwidth]{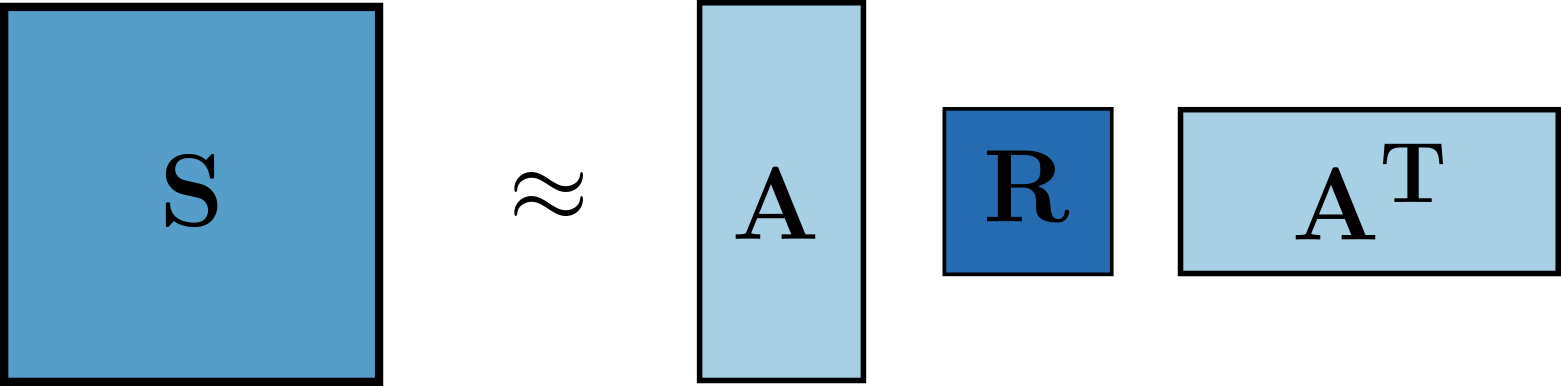}
    \caption{The DEDICOM algorithm factorizes a square matrix $\mat{S}$ into a loading matrix $\mat{A}$ and an affinity matrix $\mat{R}$.}
    \label{fig:dedicom_illustrated}
\end{figure}

DEcomposition into DIrectional COMponents (DEDICOM) is a matrix factorization technique that factorizes a square, possibly asymmetric, matrix of relationships between items into a loading matrix of low-dimensional representations of each item and an affinity matrix describing the relationships between the dimensions of the latent representation (see Figure \ref{fig:dedicom_illustrated} for an illustration).

We introduce a modified row-stochastic variation of DEDICOM, which allows for interpretable loading vectors and apply it to different matrices of word co-occurrence statistics created from Wikipedia based semi-artificial text documents. Our algorithm produces low-dimensional word embeddings, where one can interpret each latent factor as a topic that clusters words into meaningful categories. Hence, we show that row-stochastic DEDICOM successfully combines the task of learning interpretable word embeddings and extracting representative topics.

Another interesting aspect of this type of factorization is the interpretability of the affinity matrix.
An entry in the matrix directly describes the relationship between the topics of the respective row and column and one can therefore use this tool to extract topics that a certain text corpus deals with and analyse how these topics are connected in the given text.

In this work we first describe the aforementioned DEDICOM algorithm and provide details on the modified row-stochasticity constraint and on optimization.
We then present results of various experiments on semi-artificial text documents (combinations of Wikipedia articles) that show how our approach is able to capture hidden latent topics within text corpora, cluster words in a meaningful way and find relationships between these topics within the documents.

\section{Related Work}
The DEDICOM algorithm has a long history of providing interpretable matrix factorization, mostly for rather low-dimensional tasks.
First described in \cite{originalDEDICOM}, it since has been applied to analysis of social networks \cite{DEDICOMSocialNetworks}, email correspondence \cite{DEDICOMEnronEmail} and video game player behaviour \cite{Sifa2018InterpretableMF,SifaUserChurnMigration}.
DEDICOM also has successfully been employed to NLP tasks such as part of speech tagging \cite{DEDICOMPOS},
however to the best of our knowledge we provide the first implementation of DEDICOM for simultaneous word embedding learning and topic modeling.

Many works deal with the task of putting constraints on the factor matrices of the DEDICOM algorithm.
In \cite{Sifa2018InterpretableMF,DEDICOMEnronEmail}, the authors constrain the affinity matrix $\mat{R}$ to be non-negative, which aids interpretability and improves convergence behaviour if the matrix to be factorized is non-negative. 
However, their approach relies on the Kronecker product between matrices in the update step, solving a linear system of $n^2 \times k^2$, where $n$ denotes the number of items in the input matrix and $k$ the number of latent factors.
These dimensions make the application on text data, where $n$ describes the number of words in the vocabulary, a computationally futile task.
Constraints on the loading matrix, $\mat{A}$, include non-negativity as well (see \cite{DEDICOMEnronEmail}) or column-orthogonality as in \cite{Sifa2018InterpretableMF}.

In contrast, we propose a new modified row-stochasticity constraint on $\mat{A}$, which is tailored to generate interpretable word embeddings that carry semantic meaning and represent a probability distribution over latent topics.

Previous matrix factorization based methods in the NLP context mostly dealt with either word embedding learning or topic modeling, but not with both tasks combined.

For word embeddings, the GloVe \cite{glove} model factorizes an adjusted co-occurrence matrix into two matrices of the same dimension.
The work is based on a large text corpus with a vocabulary of $n\approx 400,000$ and produces word embeddings of dimension $k=300$.
In order to maximize performance on the word analogy task, the authors adjusted the co-occurrence matrix to the logarithmized co-occurrence matrix and added bias terms to the optimization objective.

A model conceived around the same time, word2vec \cite{word2vec},
calculates word embeddings not from a co-occurrence matrix but directly from the text corpus using the skip-gram or continuous-bag-of-words approach.
More recent work \cite{word2vecpmi} has shown that this construction is equivalent to matrix factorization on the pointwise mutual information (PMI) matrix of the text corpus, which makes it very similar to the glove model described above.

Both models achieve impressive results on word embedding related tasks like word analogy, however the large dimensionality of the word embeddings makes interpreting the latent factors of the embeddings impossible.

On the topic modeling side, matrix factorization methods are routinely applied as well.
Popular algorithms like non-negative matrix factorization (NMF) \cite{NMFAlgorithmsForNMF}, singular value decomposition (SVD) \cite{Furnas1988-IFU,SVD} and principal component analysis (PCA) \cite{pca} compete against the probabilistic latent dirichlet allocation (LDA) \cite{lda} to cluster the vocabulary of a word co-occurrence or document-term matrix into latent topics.\footnote{More recent expansions of these methods can be found in \cite{HellingerPCA,ImprovingTopicModels}.}  Yet, we empirically show that the implicitly learned word embeddings of these methods lack semantic meaning in terms of the cosine similarity measure.

We benchmark our approach qualitatively against these methods in Section \ref{subsec: results} and the appendix.

\section{The row-stochastic DEDICOM Model} \label{sec: dedicom}
In this section we provide a detailed theoretical view at the proposed row-stochastic DEDICOM algorithm for factorizing word co-occurrence based positive pointwise mutual information matrices.

For a given language corpus consisting of $n$ unique words $X = {x_1, \dots, x_n}$ we calculate a co-occurrence matrix $\mat{W} \in \RR^{n\times n}$ by iterating over the corpus on a word token level with a sliding context window  of specified size.
Then
\begin{equation}
    \mat{W}_{ij} = \# \text{word $i$ appears in context of word $j$}.
\end{equation}
Note that the word context window can be applied symmetrically or asymmetrically around each word. We choose a symmetric context window, which implies a symmetric co-occurrence matrix, $\mat{W}_{ij} = \mat{W}_{ji}$.

We then transform the co-occurrence matrix into the pointwise mutual information matrix (PMI),
which normalizes the counts in order to extract meaningful co-occurrences from the matrix.
Co-occurrences of words that occur regularly in the corpus are decreased since their appearance together might be nothing more than a statistical phenomenon,
the co-occurrence of words that appear less often in the corpus give us meaningful information about the relations between words and topics.
We define the PMI matrix as
\begin{equation}
    \PMI_{ij} \coloneqq \log\mat{W}_{ij} + \log N - \log N_i - \log N_j
\end{equation}
where $N \coloneqq \sum_{ij=1}^n \mat{W}_{ij}$ is the sum of all co-occurrence counts of $\mat{W}$,
$N_i \coloneqq \sum_{j=1}^n \mat{W}_{ij}$ the row sum and $N_j \coloneqq \sum_{i=1}^n \mat{W}_{ij}$ the column sum.

Since the co-occurrence matrix $\mat{W}$ is symmetrical, the transformed PMI matrix is symmetrical as well. Nevertheless, DEDICOM is able to factorize both symmetrical and non-symmetrical matrices.
We expand details on symmetrical and non-symmetrical relationships in Section \ref{subsec: interpretability}.

Additionally, we want all entries of the matrix to be non-negative,
our final matrix to be factorized is therefore the positive PMI (PPMI)
\begin{equation}
    \mat{S}_{ij} = \PPMI_{ij} = \max\{0, \PMI_{ij}\}.
\end{equation}

Our aim is to decompose this matrix using row-stochastic DEDICOM as
\begin{equation} \label{eq: objective SARAT}
    \mat{S} \approx \mat{A} \mat{R} \mat{A}^T, \quad \text{with} \quad     \mat{S}_{ij} \approx \sum_{b=1}^k\sum_{c=1}^k \mat{A}_{ib}\mat{R}_{bc}\mat{A}_{jc},
\end{equation}
where $\mat{A} \in \RR^{n \times  k}$, $\mat{R} \in \RR^{k \times  k}$, $\mat{A}^T$ denotes the transpose of $\mat{A}$ and $k \ll n$.
Literature often refers to $\mat{A}$ as the loading matrix and $\mat{R}$ as the affinity matrix. $\mat{A}$ gives us for each word $i$ in the vocabulary a vector of size $k$, the number of latent topics we wish to extract. The square matrix $\mat{R}$ then provides possibility for interpretation of the relationships between these topics.




Empirical evidence has shown that the algorithm tends to favor columns unevenly, such that a single column receives a lot more weight in its entries than the other columns.
We try to balance this behaviour by applying a column-wise z-normalization on $\mat{A}$, such that all columns have zero mean and unit variance.

In order to aid interpretability we wish each word embedding to be a distribution over all latent topics, i.e. entry $\mat{A}_{ib}$ in the word-embedding matrix provides information on how much topic $b$ describes word $i$.

To implement these constraints we therefore apply a row-wise softmax operation over the column-wise z-normalized $\mat{A}$ matrix by defining $\mat{A}' \in \RR^{n \times  k}$ as
\begin{equation} \label{def: A prime}
\begin{split}
    \mat{A}'_{ib} \coloneqq \frac{\exp(\bar{\mat{A}}_{ib})}{\sum_{b'=1}^k \exp(\bar{\mat{A}}_{ib'})}, \quad
    \bar{\mat{A}}_{ib} \coloneqq \frac{\mat{A}_{ib} - \mu_{b}}{\sigma_{b}}, \\ \quad \mu_b \coloneqq \frac{1}{n} \sum_{i=1}^n \mat{A}_{ib},
    \quad \sigma_b \coloneqq \sqrt{\frac{1}{n} \sum_{i=1}^n \left(\mat{A}_{ib} - \mu_b \right)^2}
\end{split}
\end{equation}
and optimizing $\mat{A}$ for the objective 
\begin{equation} \label{objective A prime SARAT}
    \mat{S} \approx \mat{A}' \mat{R} (\mat{A}')^T.
\end{equation}

Note that after applying the row-wise softmax operation all entries of $\mat{A}'$ are non-negative.




To judge the quality of the approximation \eqref{objective A prime SARAT} we apply the Frobenius norm, which measures the difference between $\mat{S}$ and $\mat{A}'\mat{R}\left(\mat{A}'\right)^T$.
The final loss function we optimize our model for is therefore given by
\begin{align}
    \mathcal{L}(\mat{S}, \mat{A}, \mat{R}) 
    &= \left\Vert \mat{S} - \mat{A}'\mat{R}(\mat{A}')^T  \right\Vert_F^2 \\ \label{loss}
    &= \sum_{i=1}^n \sum_{j=1}^n \left( \mat{S}_{ij} - \left(\mat{A}'\mat{R}(\mat{A}')^T\right)_{ij}\right)^2
\end{align}
with
\begin{equation}
    \left(\mat{A}'\mat{R}(\mat{A}')^T\right)_{ij} = \sum_{b=1}^k\sum_{c=1}^k \mat{A}'_{ib}\mat{R}_{bc}\mat{A}'_{jc} 
\end{equation}
and $\mat{A}'$ defined in \eqref{def: A prime}.


To optimize the loss function we train both matrices using alternating gradient descent similar to \cite{Sifa2018InterpretableMF}.
Within each optimization step we apply 
\begin{align}
    \mat{A} &\mapsfrom \mat{A} - f_{\theta}(\nabla_{\mat{A}}, \eta_{\mat{A}}), \quad \text{where} \quad \nabla_{\mat{A}} = \frac{\partial \mathcal{L}(\mat{S}, \mat{A}, \mat{R})}{\partial \mat{A}} \\
    \mat{R} &\mapsfrom \mat{R} - f_{\theta}(\nabla_{\mat{R}}, \eta_{\mat{R}}), \quad \text{where} \quad \nabla_{\mat{R}} = \frac{\partial \mathcal{L}(\mat{S}, \mat{A}, \mat{R})}{\partial \mat{R}}
\end{align}
with $\eta_{\mat{A}}, \eta_{\mat{R}} > 0$ being individual learning rates for both matrices and $f_{\theta}(\cdot)$ representing an arbitrary gradient based update rule with additional hyperparameters $\theta$.
For our experiments we employ automatic differentiation methods.
For details on the implementation of the algorithm above refer to Section \ref{subsec:training}.



\subsection{On Symmetry}\label{subsec:symmetry}
The DEDICOM algorithm is able to factorize both symmetrical and asymmetrical matrices $\mat{S}$.
For a given matrix $\mat{A}$, the symmetry of $\mat{R}$ dictates the symmetry of the product $\mat{A}\mat{R}\mat{A}^T$, since
\begin{align}
    (\mat{A}\mat{R}\mat{A}^T)_{ij} 
    &= \sum_{b=1}^k\sum_{c=1}^k \mat{A}_{ib}\mat{R}_{bc}\mat{A}_{jc}
    = \sum_{b=1}^k\sum_{c=1}^k \mat{A}_{ib}\mat{R}_{cb}\mat{A}_{jc} \\
    &= \sum_{c=1}^k\sum_{b=1}^k \mat{A}_{jc} \mat{R}_{cb} \mat{A}_{ib}
    = (\mat{A}\mat{R}\mat{A}^T)_{ji}
\end{align}
iff $\mat{R}_{cb} = \mat{R}_{bc}$ for all $b, c$.
We therefore expect a symmetric matrix $\mat{S}$ to be decomposed into $\mat{A}\mat{R}\mat{A}^T$ with a symmetric $\mat{R}$, 
which is confirmed by our experiments.
Factorizing a non-symmetric matrix leads to a non-symmetric $\mat{R}$, the asymmetric relation between items leads to asymmetric relations between the latent factors.

\subsection{On Interpretability} \label{subsec: interpretability}
We have
\begin{equation}
    \mat{S}_{ij} \approx \sum_{b=1}^k\sum_{c=1}^k \mat{A}_{ib}\mat{R}_{bc}\mat{A}_{jc},
\end{equation}
i.e. we  can estimate the probability of co-occurrence of two words $w_i$ and $w_j$ from the word embeddings $\mat{A}_i$ and $\mat{A}_j$ and the matrix $\mat{R}$, where $\mat{A_i}$ denotes the $i$-th row of $\mat{A}$.

If we want to predict the co-occurrence between words $w_i$ and $w_j$ we consider the latent topics that make up the word embeddings $\mat{A}_i$ and $\mat{A}_j$, and sum up each component from $\mat{A}_i$ with each component $\mat{A}_j$ with respect to the relationship weights given in $\mat{R}$.

\begin{figure}[ht]
    \centering
    \includegraphics[height=0.15\textwidth]{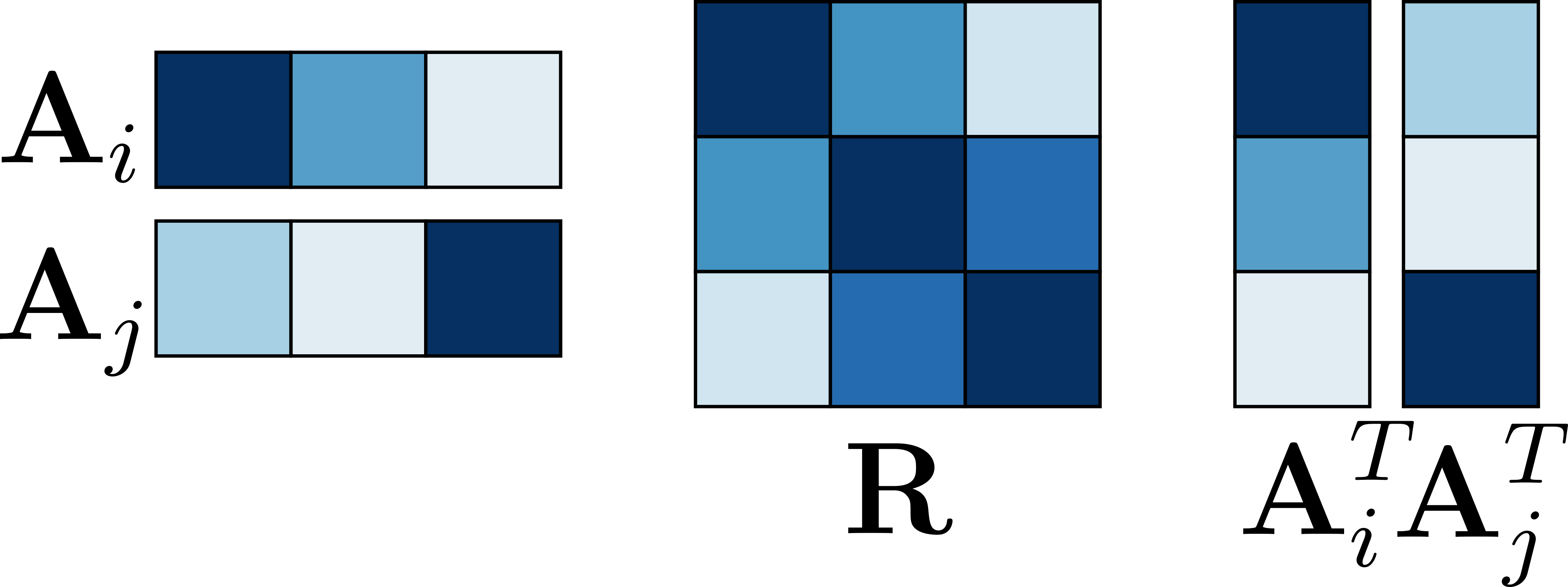}
    \caption{
        The affinity matrix $\mat{R}$ describes the relationships between the latent factors.
        Illustrated here are two word embeddings, corresponding to the words $w_i$ and $w_j$.
        Darker shades represent larger values.
        In this example we predict a large co-occurrence at $\mat{S}_{ii}$ and $\mat{S}_{jj}$ because of the large weight on the diagonal of the $\mat{R}$ matrix.
        We predict a low co-occurrence at  $\mat{S}_{ij}$ and $\mat{S}_{ji}$ since the large weights on $\mat{A}_{i1}$ and $\mat{A}_{j3}$
        interact with low weights on $\mat{R}_{13}$ and $\mat{R}_{31}$.
    }
    \label{fig: interpretability}
\end{figure}

Two words are likely to have a high co-occurrence if their word embeddings have larger weights in topics that are positively connected by the $\mat{R}$ matrix.
Likewise a negative entry $\mat{R}_{b,c}$ makes it less likely for words with high weight in the topics $b$ and $c$ to occur in the same context.
See Figure \ref{fig: interpretability} for an illustrated example.

Having an interpretable embedding model provides value beyond analysis of the affinity matrix of a single document.
The worth of word embeddings is generally measured in their usefulness for downstream tasks.
Given a prediction model based on word embeddings as one of the inputs, further analysis of the model behaviour is facilitated when latent input dimensions easily translate to semantic meaning.

In most word embedding models, the embedding vector of a single word is not particularly useful in itself.
The information only lies in its relationship (i.e. closeness or cosine similarity) to other embedding vectors.
For example, an analysis of the change of word embeddings and therefore the change of word meaning within a document corpus (for example a news article corpus) can only show how various words form different clusters or drift apart over time.
Interpretabilty of latent dimensions would provide tools to also consider the development of single words within the given topics.

\section{Experiments and Results} \label{sec: experiments_results}
In the following section we describe our experimental setup in full detail\footnote{All results are completely reproducible based on the information in this section. Our \texttt{Python} implementation to reproduce the results is available on \url{https://github.com/LarsHill/text-dedicom-paper}.} and present our results on the simultaneous topic (relation) extraction and word embedding learning task. We compare these results against competing matrix factorization methods for topic modeling, namely NMF, LDA and SVD. 

\subsection{Data} \label{subsec: data}

To conduct our experiments we leverage a synthetically created text corpus, whose documents consist of triplets of individual English Wikipedia articles. The articles are retrieved as raw text via the official Wikipedia API using the \texttt{wikipedia-api} library. Always three articles a time get concatenated to form a new artificially generated text document. 
We differentiate between thematically similar (e.g. ``Dolphin'' and ``Whale'') and thematically different articles (e.g. ``Soccer'' and ``Donald Trump'').
Each synthetic document is categorized into one of three classes: All underlying Wikipedia articles are thematically different, two articles are thematically similar and one is different, and all articles are thematically similar. Table \ref{tab:data_tab} in the appendix shows this categorization and the overall setup of our generated documents.

On each document we apply the following textual preprocessing steps. 
First, the whole document gets lower-cased. 
Second, we tokenize the text making use of the word-tokenizer from the \texttt{nltk} library and remove common English stop words, including contractions such as ``you're'' and ``we'll''. 
Lastly we clear the text from all remaining punctuation and delete digits and single characters. 

As described in Section \ref{sec: dedicom} we utilize our preprocessed document text to calculate a symmetric word co-occurrence matrix, which, after being transformed to a positive PMI matrix, functions as input and target matrix for the row-stochastic DEDICOM algorithm. 
To avoid any bias or prior information from the structure and order of the Wikipedia articles, we randomly shuffle the vocabulary before creating the co-occurrence matrix. 
When generating the matrix we only consider context words within a symmetrical window of size $7$ around the base word. Like in \cite{glove}, each context word only contributes $1/d$ to the total word pair count, given it is $d$ words apart from the base word.

The next section sheds light upon the training process of row-stochastic DEDICOM and the above mentioned competing matrix factorization methods, which will be benchmarked against our results in Section \ref{subsec: results} and in the appendix.




\subsection{Training} \label{subsec:training}

As theoretically described in Section \ref{sec: dedicom} we train row-stochastic DEDICOM with the alternating gradient descent paradigm utilizing automatic differentiation from the \texttt{PyTorch} library. 

First, we initialize the factor matrices $\mat{A} \in \RR^{n \times k}$ and $\mat{R} \in \RR^{k \times k}$, by randomly sampling all elements from a uniform distribution centered around $1$, $\mathcal{U}(0, 2)$.
Note that after applying the softmax operation on $\mat{A}$ all rows of $\mat{A}$ are stochastic.
Therefore, scaling $\mat{R}$ by
\begin{equation}
    \bar{s} \coloneqq \frac{1}{n^2}\sum_{ij}^n \mat{S}_{ij}, \label{alpha}
\end{equation}
will result in the initial decomposition $\mat{A}'\mat{R}\left(\mat{A}'\right)^T$ yielding reconstructed elements in the range of $\bar{s}$, the element mean of the PPMI matrix $\mat{S}$, and thus, speeding up convergence.

Second, $\mat{A}$ and $\mat{R}$ get iteratively updated employing the Adam optimizer \cite{kingma2014adam} with constant individual learning rates of $\eta_{\mat{A}} = 0.001$ and $\eta_{\mat{R}} = 0.01$ and hyperparameters $\beta_1 = 0.9$, $\beta_2 = 0.999$ and $\epsilon = \num{1e-8}$. 
Both learning rates were identified through an exhaustive grid search.
We train for $\texttt{num\_epochs} = 15,000$ until convergence, where each epoch consists of an alternating gradient update with respect to $\mat{A}$ and $\mat{R}$. Algorithm \ref{alg:dedicom_algorithm} illustrates the just described training procedure.

\begin{algorithm}[ht]
\caption{The row-stochastic DEDICOM algorithm}
\label{alg:dedicom_algorithm}
\begin{algorithmic}[1]
  \State initialize $\mat{A}, \mat{R} \gets U(0, 2) \cdot \bar{s}$
          \Comment{See Equation \eqref{alpha} for the definition of $\bar{s}$} 
  \State initialize $\beta_1$, $\beta_2$, $\epsilon$ \Comment{Adam algorithm hyperparameters}
  \State initialize $\eta_{\mat{A}}$, $\eta_{\mat{R}}$ \Comment{Individual learning rates}
  \Statex
  \For{$i$ in $1, \dots$, \texttt{num\_epochs}} 
  \State Calculate loss $\mathcal{L} = \mathcal{L}(\mat{S}, \mat{A}, \mat{R})$ \Comment{See Equation \eqref{loss}}
  \State $\begin{aligned}
          \mat{A} \mapsfrom \mat{A} - \text{Adam}_{\beta_1, \beta_2, \epsilon}(\nabla_{\mat{A}},  \eta_{\mat{A}}), \quad \text{where} \quad \nabla_{\mat{A}} = \frac{\partial \mathcal{L}}{\partial \mat{A}}
          \end{aligned}$
          
  \State $\begin{aligned} 
          \mat{R} \mapsfrom \mat{R} - \text{Adam}_{\beta_1, \beta_2, \epsilon}(\nabla_{\mat{R}},  \eta_{\mat{R}}), \quad \text{where} \quad \nabla_{\mat{R}} = \frac{\partial \mathcal{L}}{\partial \mat{R}}
          \end{aligned}$
  \EndFor
  \State \textbf{return} $\mat{A}'$ and $\mat{R}$, where $\mat{A}' = \text{row\_softmax}\left(\text{col\_norm}\left(\mat{A}\right)\right)$ \Comment{See Equation \eqref{def: A prime}}
\end{algorithmic}
\end{algorithm}

\begin{figure}[ht]
    \centering
    \import{figures/}{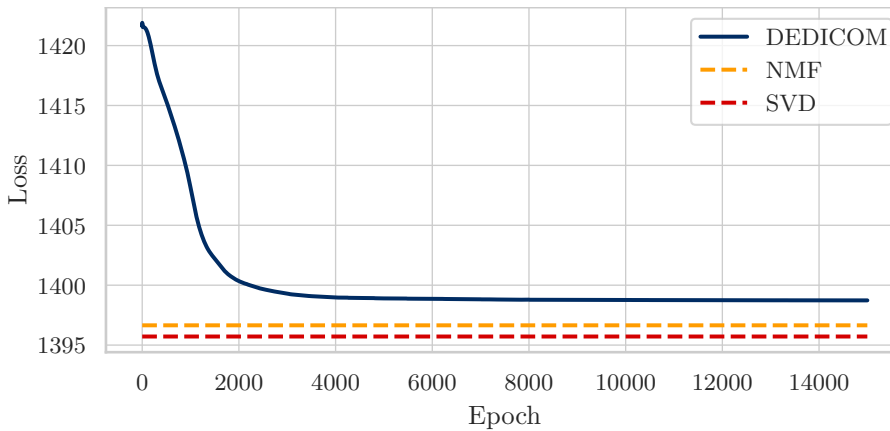}
    \caption{Reconstruction loss development during training. The $x$-axis plots the number of epochs, the $y$-axis plots the corresponding reconstruction error for each matrix factorization method.}
    \label{fig:losses}
\end{figure}

We implement NMF, LDA and SVD using the \texttt{sklearn} library. In all cases the learnable factor matrices are initialized randomly and default hyperparameters are applied during training. For NMF the multiplicative update rule from \cite{NMFAlgorithmsForNMF} is utilized. 
Figure \ref{fig:losses} shows the convergence behaviour of the row-stochastic DEDICOM training process and the final loss of NMF and SVD. 
Note that LDA optimizes a different loss function, which is why the calculated loss is not comparable and therefore excluded. 
We see that the final loss of DEDICOM locates just above the other losses, which is reasonable when considering the row stochasticity contraint on $\mat{A}$ and the reduced parameter amount of $nk + k^2$ compared to NMF ($2nk$) and SVD ($2nk + k^2$). 

\subsection{Results} \label{subsec: results} 
In the following, we present our results of training row-stochastic DEDICOM to simultaneously learn interpretable word embeddings and meaningful topic clusters and their relations. 
For compactness reasons we focus our main analysis on document id 3 in Table \ref{tab:data_tab}, ``Soccer, Bee and Johnny Depp'', and set the number of latent topics to $k=6$. 
We refer the interested reader to the appendix for results on other article combinations and comparison to other matrix factorization methods.\footnote{We provide a large scale evaluation of all article combinations listed in Table \ref{tab:data_tab}, including different choices for $k$, as supplementary material at \url{https://bit.ly/3cBxsGI}.}

\begin{figure}[!ht]
\centerfloat
  \begin{subfigure}[b]{0.22\textwidth}
     \includegraphics[page=1, width=\linewidth]{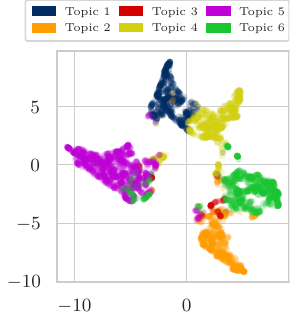}
    \caption{}
    \label{fig:scatter_topics_dedicom_soccer_bee_depp}
  \end{subfigure}
  \begin{subfigure}[b]{0.22\textwidth}
    \includegraphics[page=2, width=\linewidth]{dedicom_pgfplots.pdf}
    \caption{}
    \label{fig:scatter_titles_dedicom_soccer_bee_depp}
  \end{subfigure}
  \begin{subfigure}[b]{0.26\textwidth}
    \includegraphics[page=3, width=\linewidth]{dedicom_pgfplots.pdf}
    \caption{}
    \label{fig:heatmap_dedicom_soccer_bee_depp}
  \end{subfigure}
 \caption{(a) 2-dimensional representation of word embeddings $\mat{A}'$ colored by topic assignment. (b) 2-dimensional representation of word embeddings $\mat{A}'$ colored by original Wikipedia article assignment (words that occur in more than one article are excluded). (c) Colored heatmap of affinity matrix $\mat{R}$.}
 \vspace{1.2em}
\begin{tabular}{lllllll}
\toprule
&  Topic 1 &    Topic 2 &    Topic 3 &      Topic 4 &   Topic 5 &       Topic 6 \\
 & \#619 &     \#1238 &      \#628 &      \#595 &      \#612 &      \#389 \\
\midrule
\multirow{2}{*}{1} &   ball &       film &    salazar &          cup &      bees &         heard \\
 &  (0.77) &    (0.857) &    (0.201) &      (0.792) &   (0.851) &       (0.738) \\
\multirow{2}{*}{2} & penalty &    starred &   geoffrey &     football &   species &         court \\
 & (0.708) &    (0.613) &      (0.2) &      (0.745) &   (0.771) &       (0.512) \\
\multirow{2}{*}{3} &     may &       role &       rush &         fifa &       bee &          depp \\
&  (0.703) &    (0.577) &      (0.2) &      (0.731) &   (0.753) &       (0.505) \\
\multirow{2}{*}{4} & referee &     series &    brenton &        world &    pollen &       divorce \\
 & (0.667) &    (0.504) &    (0.199) &      (0.713) &   (0.658) &       (0.454) \\
\multirow{2}{*}{5} &    goal &     burton &  hardwicke &     national &     honey &       alcohol \\
 &  (0.66) &    (0.492) &    (0.198) &      (0.639) &   (0.602) &       (0.435) \\
\multirow{2}{*}{6} &    team &  character &   thwaites &         uefa &   insects &       paradis \\
 & (0.651) &    (0.465) &    (0.198) &      (0.623) &   (0.576) &        (0.42) \\
\multirow{2}{*}{7} &  players &     played &  catherine &  continental &      food &  relationship \\
&  (0.643) &    (0.451) &    (0.198) &      (0.582) &   (0.536) &       (0.419) \\
\multirow{2}{*}{8} &   player &   director &       kaya &        teams &     nests &         abuse \\
&  (0.639) &     (0.45) &    (0.198) &      (0.576) &   (0.529) &        (0.41) \\
\multirow{2}{*}{9} &     play &    success &      melfi &     european &  solitary &       stating \\
&  (0.606) &    (0.438) &    (0.198) &       (0.57) &   (0.513) &       (0.408) \\
\multirow{2}{*}{10} &     game &       jack &      raimi &  association &  eusocial &        stated \\
&  (0.591) &    (0.434) &    (0.198) &      (0.563) &   (0.505) &       (0.402) \\
\bottomrule
\end{tabular}
\captionof{table}{Each column lists the top 10 representative words per dimension of the basis matrix $\mat{A}'$.}
\label{topics_dedicom}

\end{figure}

In a first step, we evaluate the quality of the learned latent topics by assigning each word embedding $\mat{A}'_i \in \RR^{1 \times k}$ to the latent topic dimension that represents the maximum value in $\mat{A}'_i$, i.e.
\begin{align*}
    \mat{A}'_i &= \begin{bmatrix} 0.05 & 0.03 & 0.02 & 0.14 & 0.70 & 0.06 \end{bmatrix} \\
    \text{argmax}\left( \mat{A}'_i \right) &= 5,
\end{align*}
and thus, $\mat{A}'_i$ gets matched to Topic 5. Next, we decreasingly sort the words within each topic based on their matched topic probability. Table \ref{topics_dedicom} shows the overall number of allocated words and the resulting top 10 words per topic together with each matched probability.

\begin{table}[!ht]
\footnotesize
\centering
\begin{tabular}{lllllll}
\toprule
& Topic 1 &  Topic 2 &   Topic 3 &   Topic 4 &  Topic 5 &     Topic 6 \\
\midrule
\multirow{2}{*}{0} &    ball &     film &   salazar &       cup &     bees &       heard \\
 &   (1.0) &    (1.0) &     (1.0) &     (1.0) &    (1.0) &       (1.0) \\
\multirow{2}{*}{1} & penalty &  starred &  geoffrey &      fifa &      bee &       court \\
 & (0.994) &  (0.978) &     (1.0) &   (0.995) &  (0.996) &     (0.966) \\
\multirow{2}{*}{2} & referee &     role &      rush &  national &  species &     divorce \\
 & (0.992) &  (0.964) &     (1.0) &   (0.991) &  (0.995) &     (0.944) \\
\multirow{2}{*}{3} &     may &   burton &    bardem &     world &   pollen &     alcohol \\
 & (0.989) &  (0.937) &     (1.0) &   (0.988) &  (0.986) &     (0.933) \\
\multirow{2}{*}{4} &    goal &   series &   brenton &      uefa &    honey &       abuse \\
 & (0.986) &  (0.935) &     (1.0) &   (0.987) &  (0.971) &     (0.914) \\
 \midrule
\multirow{2}{*}{0} & penalty &  starred &  geoffrey &  football &  species &       court \\
 &   (1.0) &    (1.0) &     (1.0) &     (1.0) &    (1.0) &       (1.0) \\
\multirow{2}{*}{1} & referee &     role &      rush &      fifa &     bees &     divorce \\
 & (0.999) &  (0.994) &     (1.0) &   (0.994) &  (0.995) &     (0.995) \\
\multirow{2}{*}{2} &    goal &   series &   salazar &  national &      bee &     alcohol \\
 & (0.998) &  (0.985) &     (1.0) &   (0.983) &   (0.99) &     (0.987) \\
\multirow{2}{*}{3} &  player &   burton &   brenton &       cup &   pollen &       abuse \\
 & (0.997) &  (0.981) &     (1.0) &   (0.983) &   (0.99) &     (0.982) \\
\multirow{2}{*}{4} &    ball &     film &  thwaites &     world &  insects &  settlement \\
 & (0.994) &  (0.978) &     (1.0) &   (0.982) &  (0.977) &     (0.978) \\
\bottomrule
\end{tabular}
\caption{For the most significant two words per topic, the four nearest neighbors based on cosine similarity are listed.}
\label{similar_words_dedicom}
\end{table}

Indicated by the high assignment probabilities, one can see that columns 1, 2, 4, 5 and 6 represent distinct topics, which easily can be interpreted. 
Topic 1 and 4 are related to soccer, where 1 focuses on the game mechanics and 4 on the organisational and professional aspect of the game. 
Topic 2 and 6 clearly refer to Johnny Depp, where 2 focuses on his acting career and 6 on his difficult relationship to Amber Heard. 
The fifth topic obviously relates to the insect ``bee''. 
In contrast, Topic 3 does not allow for any interpretation and all assignment probabilities are significantly lower than for the other topics.

Further, we analyze the relations between the topics by visualizing the trained $\mat{R}$ matrix as a heatmap (see Figure \ref{fig:heatmap_dedicom_soccer_bee_depp}).

One thing to note is the symmetry of $\mat{R}$ which is a first indicator of a successful reconstruction, $\mat{A}'\mat{R}\left(\mat{A}'\right)^T$, (see Section \ref{subsec:symmetry}). 
Also, the main diagonal elements are consistently blue (positive), which suggests a high distinction between the topics. 
Although not very strong one can still see a connection between Topic 2 and 6 indicated by the light blue entry $\mat{R}_{26} = \mat{R}_{62}$. 
While the suggested relation between Topic 1 and 4 is not clearly visible, element $\mat{R}_{14} = \mat{R}_{41}$ is the least negative one for Topic 1.
In order to visualize the topic cluster quality we utilize UMAP (Uniform Manifold Approximation ad Projection) \cite{umap} to map the $k$-dimensional word embeddings to a 2-dimensional space. 
Figure \ref{fig:scatter_topics_dedicom_soccer_bee_depp} illustrates this low-dimensional representation of $\mat{A}'$, where each word is colored based on the above described word to topic assignment.
In conjunction with Table \ref{topics_dedicom} one can nicely see that Topic 2 and 6 (Johnny Depp) and Topic 1 and 4 (Soccer) are close to each other. 
Hence, Figure \ref{fig:scatter_topics_dedicom_soccer_bee_depp} implicitly shows the learned topic relations as well and arguably better than $\mat{R}$.

As an additional benchmark, Figure \ref{fig:scatter_titles_dedicom_soccer_bee_depp} plots the same 2-dimensional representation, but now each word is colored based on the original Wikipedia article it belonged to. Words that occur in more than one article are not considered in this plot.

Directly comparing Figure \ref{fig:scatter_titles_dedicom_soccer_bee_depp} and \ref{fig:scatter_topics_dedicom_soccer_bee_depp} shows that row-stochastic DEDICOM does not only recover the original articles but also finds entirely new topics, which in this case represent subtopics of the articles.
Let us emphasize that for all thematically similar article combinations, the found topics are usually not subtopics of a single article, but rather novel topics that might span across multiple Wikipedia articles (see for example Table \ref{table:dolphin_shark_whale_dedicom} in the appendix).
As mentioned at the top of this section, we are not only interested in learning meaningful topic clusters, but also in training interpretable word embeddings that capture semantic meaning. 

Hence, we select within each topic the two most representative words and calculate the cosine similarity between their word embeddings and all other word embeddings stored in $\mat{A}'$. Table \ref{similar_words_dedicom} shows the 4 nearest neighbors based on cosine similarity for the top 2 words in each topic.
We observe a high thematical similarity between words with large cosine similarity, indicating the usefulness of the rows of $\mat{A}'$ as word embeddings.

In comparison to DEDICOM, other matrix factorization methods also provide a useful clustering of words into topics, with varying degree of granularity and clarity. 
However, the application of these methods as word embedding algorithms mostly fails on the word similarity task, with words close in cosine similarity seldom sharing the same thematical similarity we have seen in DEDICOM. This can be seen in Table \ref{table: soccer_bee_depp_nmf}, which shows for each method, NMF, LDA and SVD, the resulting word to topic clustering and the cosine nearest neighbors of the top two word embeddings per topic. While the individual topics extracted by NMF look very reasonable, its word embeddings do not seem to carry any semantic meaning based on cosine similarity; e.g. the four nearest neighbors of ``ball'' are ``invoke'', ``replaced'', ``scores'' and ``subdivided''. A similar nonsensical picture can be observed for the other main topic words. LDA and SVD perform slightly better on the similar word task, although not all similar words appear to be sensible, e.g. ``children'', ``detective'', ``crime'', ``magazine'' and ``barber''. Also, some topics cannot be clearly defined due to mixed word assignments, e.g. Topic 4 for LDA and Topic 1 for SVD.

For a comprehensive overview of our results for other article combinations, we refer to Tables \ref{table:dolphin_shark_whale_dedicom}, \ref{table: dolphin_shark_whale_nmf}, \ref{table: soccer_tennis_rugby_dedicom}, \ref{table: soccer_tennis_rugby_nmf} and Figures \ref{fig:plots_dolphin_shark_whale}, \ref{fig:plots_soccer_tennis_rugby} in the Appendix.


\section{Conclusion and Outlook}

We propose a row-stochasticity constrained version of the DEDICOM algorithm that is able to factorize the pointwise mutual information matrices of text documents into meaningful topic clusters all the while providing interpretable word embeddings for each vocabulary item.
Our study on semi-artificial data from Wikipedia articles has shown that this method recovers the underlying structure of the text corpus and provides topics with thematic granularity,
meaning the extracted latent topics are more specific than a simple clustering of articles.
A comparison to related matrix factorization methods has shown that the combination of topic modeling and interpretable word embedding learning given by our algorithm is unique in its class.

In future work we will expand on the idea of comparing topic relationships between multiple documents, possibly over time, with individual co-occurrence matrices resulting in stacked topic relationship matrices but shared word embeddings. Further extending this notion, we plan to utilize time series analysis to discover temporal relations between extracted topics and to potentially identify trends.

\section{Acknowledgement}

The authors of this work were supported by the Competence Center for Machine Learning Rhine Ruhr (ML2R) which is funded by the Federal Ministry of Education and Research of Germany (grant no. 01|S18038C). We gratefully acknowledge this support.




%
%
%

\bibliographystyle{splncs04}
\bibliography{dedicom}

\begin{thebibliography}{10}
\providecommand{\url}[1]{\texttt{#1}}
\providecommand{\urlprefix}{URL }
\providecommand{\doi}[1]{https://doi.org/#1}

\bibitem{DEDICOMSocialNetworks}
Andrzej, A.H., Cichocki, A., Dinh, T.V.: Nonnegative dedicom based on tensor decompositions for social networks exploration. Aust. J. Intell. Inf. Process. Syst.  \textbf{12} (2010)

\bibitem{DEDICOMEnronEmail}
Bader, B.W., Harshman, R.A., Kolda, T.G.: Pattern analysis of directed graphs using dedicom: an application to enron email. Office of Scientific \& Technical Information Technical Reports  (2006)

\bibitem{lda}
Blei, D.M., Ng, A.Y., Jordan, M.I.: Latent dirichlet allocation. J. Mach. Learn. Res.  \textbf{3},  993–--1022 (2003)

\bibitem{DEDICOMPOS}
Chew, P., Bader, B., Rozovskaya, A.: Using {DEDICOM} for completely unsupervised part-of-speech tagging. In: Proceedings of the Workshop on Unsupervised and Minimally Supervised Learning of Lexical Semantics. pp. 54--62. Association for Computational Linguistics, Boulder, Colorado, USA (2009)

\bibitem{Furnas1988-IFU}
Furnas, G.W., Deerwester, S., Dumais, S.T., Landauer, T.K., Harshman, R.A., Streeter, L.A., Lochbaum, K.E.: {Information Retrieval Using A Singular Value Decomposition Model of Latent Semantic Structure}. In: Proc. of ACM SIGIR (1988)

\bibitem{originalDEDICOM}
Harshman, R., Green, P., Wind, Y., Lundy, M.: A model for the analysis of asymmetric data in marketing research. Marketing Science  \textbf{1},  205--242 (1982)

\bibitem{pca}
Jolliffe, I.: Principal component analysis. John Wiley and Sons Ltd (2005)

\bibitem{kingma2014adam}
Kingma, D.P., Ba, J.: Adam: A method for stochastic optimization. arXiv preprint arXiv:1412.6980  (2014)

\bibitem{HellingerPCA}
Lebret, R., Collobert, R.: Word embeddings through hellinger {PCA}. In: Proceedings of the 14th Conference of the {E}uropean Chapter of the Association for Computational Linguistics. pp. 482--490. Association for Computational Linguistics, Gothenburg, Sweden (2014)

\bibitem{NMFAlgorithmsForNMF}
Lee, D.D., Seung, H.S.: Algorithms for non-negative matrix factorization. In: Proceedings of the 13th International Conference on Neural Information Processing Systems. pp. 535–--541. NIPS’00, MIT Press, Cambridge, MA, USA (2000)

\bibitem{word2vecpmi}
Levy, O., Goldberg, Y.: Neural word embedding as implicit matrix factorization. In: Proceedings of the 27th International Conference on Neural Information Processing Systems - Volume 2. pp. 2177--2185. NIPS’14, MIT Press, Cambridge, MA, USA (2014)

\bibitem{umap}
McInnes, L., Healy, J., Melville, J.: Umap: Uniform manifold approximation and projection for dimension reduction (2018)

\bibitem{word2vec}
Mikolov, T., Sutskever, I., Chen, K., Corrado, G., Dean, J.: Distributed representations of words and phrases and their compositionality (2013)

\bibitem{ImprovingTopicModels}
Nguyen, D.Q., Billingsley, R., Du, L., Johnson, M.: Improving topic models with latent feature word representations. Transactions of the Association for Computational Linguistics  \textbf{3}(0) (2015)

\bibitem{glove}
Pennington, J., Socher, R., Manning, C.: {G}love: Global vectors for word representation. In: Proceedings of the 2014 Conference on Empirical Methods in Natural Language Processing ({EMNLP}). pp. 1532--1543. Association for Computational Linguistics, Doha, Qatar (2014)

\bibitem{Sifa2018InterpretableMF}
Sifa, R., Ojeda, Cvejoski, K., Bauckhage, C.: Interpretable matrix factorization with stochasticity constrained nonnegative dedicom. In: Proceedings of KDML-LWDA (2018)

\bibitem{SifaUserChurnMigration}
Sifa, R., Ojeda, C., Bauckage, C.: User churn migration analysis with dedicom. In: Proceedings of the 9th ACM Conference on Recommender Systems. pp. 321--–324. RecSys ’15, Association for Computing Machinery, New York, NY, USA (2015)

\bibitem{SVD}
{YongchangWang}, {Zhu}, L.: Research and implementation of svd in machine learning. In: 2017 IEEE/ACIS 16th International Conference on Computer and Information Science (ICIS). pp. 471--475 (2017)

\end{thebibliography}
\clearpage

\appendix

\section*{Appendix}
\begin{table}[h]
\footnotesize
\centering

\begin{tabular}{cllll}

\toprule
ID & Selection Type                       &     Article 1       &       Article 2         &      Article 3          \\
\midrule
1 & \multirow{4}{*}{different}      &     Donald Trump    &       New York City     &      Shark              \\
2 &                                     &     Shark           &       Bee               &      Elephant           \\
3 &                                     &     Soccer          &       Bee               &      Johnny Depp        \\
4 &                                     &     Tennis          &       Dolphin           &      New York City      \\
\midrule
5 & \multirow{4}{*}{mixed}       &     Donald Trump    &       New York City     &      Michael Bloomberg  \\
6 &                                      &     Soccer          &       Tennis            &      Boxing             \\
7 &                                      &     Brad Pitt       &       Leonardo Dicaprio &      Rafael Nadal       \\
8 &                                      &     Apple (company) &       Google            &      Walmart            \\
\midrule
9 & \multirow{4}{*}{similar}             &     Shark           &       Dolphin           &      Whale              \\
10 &                                      &     Germany         &       Belgium           &      France             \\
11 &                                      &     Soccer          &       Tennis            &      Rugby football     \\
12 &                                     &     Apple (company) &       Google            &      Amazon (company)   \\
\bottomrule
\end{tabular}
\caption{Overview of our semi-artifical dataset. Each synthetic sample consists of the corresponding Wikipedia articles 1 -- 3. 
We differentiate between \emph{different} articles, i.e. articles that have little thematical overlap (for example a person and a city, a fish and an insect or a ball game and a combat sport), and \emph{similar} articles, i.e. articles with large thematical overlap (for example European countries, tech companies or aquatic animals).
We group our dataset into different samples (3 articles that are pairwise different), similar samples (3 articles that are all similar) and mixed samples (2 similar articles, 1 different).}
\label{tab:data_tab}
\end{table}

\clearpage


\begin{table}[!ht]
    \centering
    \caption*{Articles: ``Soccer'', ``Bee'', ``Johnny Depp''}
    \resizebox{0.75\textwidth}{!}{%
    \begin{tabular}{llllllll}
        \toprule
        & & Topic 1 & Topic 2 & Topic 3 & Topic 4 & Topic 5 & Topic 6 \\
        \midrule[1pt]
        & & \#619 &     \#1238 &      \#628 &      \#595 &      \#612 &      \#389 \\
        \multirow{20}{*}{NMF} & 1 & ball & bees & film & football & heard & album \\
        & 2 & may & species & starred & cup & depp & band \\
        & 3 & penalty & bee & role & world & court & guitar \\
        & 4 & referee & pollen & series & fifa & alcohol & vampires \\
        & 5 & players & honey & burton & national & relationship & rock \\
        & 6 & team & insects & character & association & stated & hollywood \\
        & 7 & goal & food & films & international & divorce & song \\
        & 8 & game & nests & box & women & abuse & released \\
        & 9 & player & solitary & office & teams & paradis & perry \\
        & 10 & play & eusocial & jack & uefa & stating & debut \\
        \cmidrule[1pt]{2-8}

        & 0 & ball & bees & film & football & heard & album \\
        & 1 & invoke & odors & burtondirected & athenaeus & crew & jones \\
        & 2 & replaced & tufts & tone & paralympic & alleging & marilyn \\
        & 3 & scores & colour & landau & governing & oped & roots \\
        & 4 & subdivided & affected & brother & varieties & asserted & drums \\
        \cmidrule{2-8}
        & 0 & may & species & starred & cup & depp & band \\
        & 1 & yd & niko & shared & inaugurated & refer & heroes \\
        & 2 & ineffectiveness & commercially & whitaker & confederation & york & bowie \\
        & 3 & tactical & microbiota & eccentric & gold & leaders & debut \\
        & 4 & slower & strategies & befriends & headquarters & nonindian & solo \\
        \midrule[1pt]

        & & \#577 &      \#728 &      \#692 &      \#607 &      \#663 &      \#814 \\
        \multirow{20}{*}{LDA} & 1 & film & football & depp & penalty & bees & species \\
        & 2 & series & women & children & heard & flowers & workers \\
        & 3 & man & association & life & ball & bee & solitary \\
        & 4 & played & fifa & role & direct & honey & players \\
        & 5 & pirates & teams & starred & referee & pollen & colonies \\
        & 6 & character & games & alongside & red & food & eusocial \\
        & 7 & along & world & actor & time & increased & nest \\
        & 8 & cast & cup & stated & goal & pollination & may \\
        & 9 & also & game & burton & scored & times & size \\
        & 10& hollow & international & playing & player & larvae & egg \\
        \cmidrule[1pt]{2-8}

        & 0 & film & football & depp & penalty & bees & species \\
        & 1 & charlie & cup & critical & extra & bee & social \\
        & 2 & near & canada & february & kicks & insects & chosen \\
        & 3 & thinking & zealand & script & inner & authors & females \\
        & 4 & shadows & activities & song & moving & hives & subspecies \\
        \cmidrule{2-8}
        & 0 & series & women & children & heard & flowers & workers \\
        & 1 & crybaby & fifa & detective & allison & always & carcases \\
        & 2 & waters & opera & crime & serious & eusociality & lived \\
        & 3 & sang & exceeding & magazine & allergic & varroa & provisioned \\
        & 4 & cast & cuju & barber & cost & wing & cuckoo \\
        \midrule[1pt]

        & & \#1228 &      \#797 &      \#628 &      \#369 &      \#622 &      \#437 \\
        \multirow{20}{*}{SVD} & 1 & bees & depp & game & cup & heard & beekeeping \\
        & 2 & also & film & ball & football & court & increased \\
        & 3 & bee & starred & team & fifa & divorce & honey \\
        & 4 & species & role & players & world & stating & described \\
        & 5 & played & series & penalty & european & alcohol & use \\
        & 6 & time & burton & play & uefa & paradis & wild \\
        & 7 & one & character & may & national & documents & varroa \\
        & 8 & first & actor & referee & europe & abuse & mites \\
        & 9 & two & released & competitions & continental & settlement & colony \\
        & 10& pollen & release & laws & confederation & sued & flowers \\
        \cmidrule[1pt]{2-8}

        & 0 & bees & depp & game & cup & heard & beekeeping \\
        & 1 & bee & iii & correct & continental & alleging & varroa \\
        & 2 & develops & racism & abandoned & contested & attempting & animals \\
        & 3 & studied & appropriation & maximum & confederations & finalized & mites \\
        & 4 & crops & march & clear & conmebol & submitted & plato \\
        \cmidrule{2-8}
        & 0 & also & film & ball & football & court & increased \\
        & 1 & although & waters & finely & er & declaration & usage \\
        & 2 & told & robinson & poised & suffix & issued & farmers \\
        & 3 & chosen & scott & worn & word & restraining & mentioned \\
        & 4 & stars & costars & manner & appended & verbally & aeneid \\
        \bottomrule

    \end{tabular}}
    \caption{For each evaluated matrix factorization method we display the top 10 words for each topic and the 5 most similar words based on cosine similarity for the 2 top words from each topic.}
    \label{table: soccer_bee_depp_nmf}
\end{table}
\clearpage

\begin{figure}[!ht]
    \begin{minipage}[b]{\linewidth}
    \centering
    \caption*{Articles: ``Dolphin'', ``Shark'', ``Whale''}
    \resizebox{0.7\textwidth}{!}{%
    \begin{tabular}{lllllll}
        \toprule
        & Topic 1 & Topic 2 & Topic 3 & Topic 4 & Topic 5 & Topic 6 \\
        & \#460 &      \#665 &      \#801 &      \#753 &      \#854 &      \#721 \\
        \midrule[1pt]
        \multirow{2}{*}{1} & shark & calf & ship & conservation & water & dolphin \\
         & (0.665) &     (0.428) &     (0.459) &       (0.334) &   (0.416) &     (0.691) \\
        \multirow{2}{*}{2} & sharks & months & became & countries & similar & dolphins \\
         & (0.645) &     (0.407) &     (0.448) &       (0.312) &   (0.374) &     (0.655) \\
        \multirow{2}{*}{3} & fins & calves & poseidon & government & tissue & captivity \\
         & (0.487) &     (0.407) &      (0.44) &       (0.309) &   (0.373) &     (0.549) \\
         \multirow{2}{*}{4} & killed & females & riding & wales & body & wild \\
         & (0.454) &     (0.399) &     (0.426) &       (0.304) &   (0.365) &     (0.467) \\
         \multirow{2}{*}{5} & million & blubber & dionysus & bycatch & swimming & behavior \\
         & (0.451) &     (0.374) &     (0.422) &        (0.29) &   (0.357) &     (0.461) \\
         \multirow{2}{*}{6} & fish & young & ancient & cancelled & blood & bottlenose \\
         & (0.448) &      (0.37) &      (0.42) &       (0.288) &   (0.346) &     (0.453) \\
         \multirow{2}{*}{7} & international & sperm & deity & eastern & surface & sometimes \\
         & (0.442) &     (0.356) &     (0.412) &       (0.287) &   (0.344) &     (0.449) \\
         \multirow{2}{*}{8} & fin & born & ago & policy & oxygen & human \\
         & (0.421) &     (0.355) &     (0.398) &       (0.286) &    (0.34) &     (0.421) \\
         \multirow{2}{*}{9} & fishing & feed & melicertes & control & system & less \\
         & (0.405) &     (0.349) &     (0.395) &       (0.285) &   (0.336) &      (0.42) \\
         \multirow{2}{*}{10} & teeth & mysticetes & greeks & imminent & swim & various \\
         & (0.398) &     (0.341) &     (0.394) &       (0.282) &   (0.336) &     (0.418) \\
        \midrule[1pt]
         \multirow{2}{*}{0} & shark & calf & ship & conservation & water & dolphin \\
         & (1.0) &    (1.0) &     (1.0) &         (1.0) &     (1.0) &       (1.0) \\
         \multirow{2}{*}{2} & sharks & calves & dionysus & south & prey & dolphins \\
         & (0.981) &  (0.978) &   (0.995) &       (0.981) &   (0.964) &     (0.925) \\
         \multirow{2}{*}{3} & fins & females & riding & states & swimming & sometimes \\
         & (0.958) &  (0.976) &   (0.992) &       (0.981) &   (0.959) &     (0.909) \\
         \multirow{2}{*}{4} & killed & months & deity & united & allows & another \\
         & (0.929) &  (0.955) &   (0.992) &       (0.978) &   (0.957) &     (0.904) \\
         \multirow{2}{*}{5} & fishing & young & poseidon & endangered & swim & bottlenose \\
         & (0.916) &  (0.948) &   (0.987) &       (0.976) &   (0.947) &     (0.903) \\
        \midrule
         \multirow{2}{*}{0} & sharks & months & became & countries & similar & dolphins \\
         & (1.0) &    (1.0) &     (1.0) &         (1.0) &     (1.0) &       (1.0) \\
         \multirow{2}{*}{2} & shark & born & old & eastern & surface & behavior \\
         & (0.981) &  (0.992) &   (0.953) &       (0.991) &   (0.992) &     (0.956) \\
         \multirow{2}{*}{3} & fins & young & later & united & brain & sometimes \\
         & (0.936) &  (0.992) &   (0.946) &       (0.989) &    (0.97) &     (0.945) \\
         \multirow{2}{*}{4} & tiger & sperm & ago & caught & sound & various \\
         & (0.894) &  (0.985) &   (0.939) &       (0.987) &   (0.968) &     (0.943) \\
         \multirow{2}{*}{5} & killed & calves & modern & south & object & less \\
         & (0.887) &  (0.984) &   (0.937) &       (0.979) &   (0.965) &     (0.937) \\
         \bottomrule
    \end{tabular}}
    \caption{Top half lists the top 10 representative words per dimension of the basis matrix $A$, bottom half lists the 5 most similar words based on cosine similarity for the 2 top words from each topic.}

        \captionof{table}{Top half lists the top 10 representative words per dimension of the basis matrix $A$, bottom half lists the 5 most similar words based on cosine similarity for the 2 top words from each topic.}
        \label{table:dolphin_shark_whale_dedicom}
    \end{minipage}
    \hspace{1cm}
    \begin{minipage}[b]{\linewidth}
        \centerfloat
          \begin{subfigure}[b]{0.18\textwidth}
             \includegraphics[page=4, width=\linewidth]{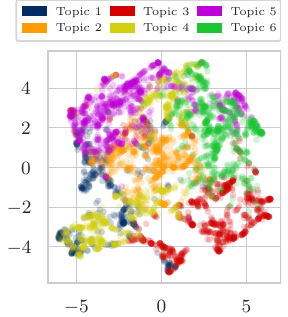}
            \caption{}
            \label{fig:scatter_topics_dedicom_shark_dolphin_whale}
          \end{subfigure}
          \begin{subfigure}[b]{0.18\textwidth}
            \includegraphics[page=5, width=\linewidth]{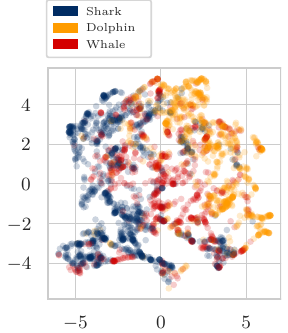}
            \caption{}
            \label{fig:scatter_titles_dedicom_shark_dolphin_whale}
          \end{subfigure}
          \begin{subfigure}[b]{0.22\textwidth}
            \includegraphics[page=6, width=\linewidth]{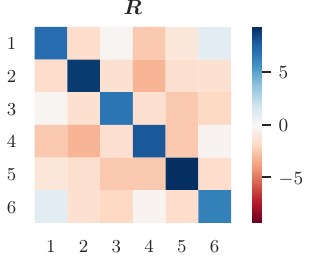}
            \caption{}
            \label{fig:heatmap_dedicom_shark_dolphin_whale}
          \end{subfigure}
         \caption{(a) 2-dimensional representation of word embeddings $\mat{A}'$ colored by topic assignment. (b) 2-dimensional representation of word embeddings $\mat{A}'$ colored by original Wikipedia article assignment (words that occur in more than one article are excluded). (c) Colored heatmap of affinity matrix $\mat{R}$.}
         \label{fig:plots_dolphin_shark_whale}
    \end{minipage}
\end{figure}
\clearpage

\begin{table}[!ht]
    \scriptsize
    \centering

    \caption*{Articles: ``Dolphin'', ``Shark'', ``Whale''}
    \resizebox{0.84\textwidth}{!}{%
    \begin{tabular}{llllllll}
        \toprule
        & & Topic 1 & Topic 2 & Topic 3 & Topic 4 & Topic 5 & Topic 6 \\
        \midrule[1pt]
        & & \#492 &      \#907 &      \#452 &      \#854 &      \#911 &      \#638 \\
        \multirow{20}{*}{NMF} & 1 & blood & international & evidence & sonar & ago & calf \\
            & 2 & body & killed & selfawareness & may & teeth & young \\
            & 3 & heart & states & ship & surface & million & females \\
            & 4 & gills & conservation & dionysus & clicks & mysticetes & captivity \\
            & 5 & bony & new & came & prey & whales & calves \\
            & 6 & oxygen & united & another & use & years & months \\
            & 7 & organs & shark & important & underwater & baleen & born \\
            & 8 & tissue & world & poseidon & sounds & cetaceans & species \\
            & 9 & water & endangered & mark & known & modern & male \\
            & 10 & via & islands & riding & similar & extinct & female \\
            \cmidrule[1pt]{2-8}

            & 0 & blood & international & evidence & sonar & ago & calf \\
            & 1 & travels & proposal & flaws & poisoned & consist & uninformed \\
            & 2 & enters & lipotidae & methodological & signals & specialize & primary \\
            & 3 & vibration & banned & nictating & – & legs & born \\
            & 4 & tolerant & iniidae & wake & emitted & closest & leaner \\
            \cmidrule{2-8}
            & 0 & body & killed & selfawareness & may & teeth & young \\
            & 1 & crystal & law & legendary & individuals & fuel & brood \\
            & 2 & blocks & consumers & humankind & helping & lamp & lacking \\
            & 3 & modified & pontoporiidae & helpers & waste & filterfeeding & accurate \\
            & 4 & slits & org & performing & depression & krill & consistency \\
            \midrule[1pt]

            & & \#650 &      \#785 &      \#695 &      \#815 &      \#635 &      \#674 \\
        \multirow{20}{*}{LDA} & 1 & killed & teeth & head & species & meat & air \\
            & 2 & system & baleen & fish & male & whale & using \\
            & 3 & endangered & mysticetes & dolphin & females & ft & causing \\
            & 4 & often & ago & fin & whales & fisheries & currents \\
            & 5 & close & jaw & eyes & sometimes & also & sounds \\
            & 6 & sharks & family & fat & captivity & ocean & groups \\
            & 7 & countries & water & navy & young & threats & sound \\
            & 8 & since & includes & popular & shark & children & research \\
            & 9 & called & allow & tissue & female & population & clicks \\
            & 10& vessels & greater & tail & wild & bottom & burst \\
            \cmidrule[1pt]{2-8}

            & 0 & killed & teeth & head & species & meat & air \\
            & 1 & postures & dense & underside & along & porbeagle & australis \\
            & 2 & dolphinariums & cetacea & grooves & another & source & submerged \\
            & 3 & town & tourism & eyesight & long & activities & melbourne \\
            & 4 & onethird & planktonfeeders & osmoregulation & sleep & comparable & spear \\

            \cmidrule{2-8}
            & 0 & system & baleen & fish & male & whale & using \\
            & 1 & dominate & mysticetes & mostly & females & live & communication \\
            & 2 & close & distinguishing & swim & aorta & human & become \\
            & 3 & controversy & unique & due & female & cold & associated \\
            & 4 & agree & remove & whole & position & parts & mirror \\

            \midrule[1pt]

            & & \#1486 &      \#544 &      \#605 &      \#469 &      \#539 &      \#611 \\
        \multirow{20}{*}{SVD} & 1 & dolphins & water & shark & million & poseidon & dolphin \\
            & 2 & species & body & sharks & years & became & meat \\
            & 3 & whales & tail & fins & ago & ship & family \\
            & 4 & fish & teeth & international & whale & riding & river \\
            & 5 & also & flippers & killed & two & evidence & similar \\
            & 6 & large & tissue & fishing & calf & melicertes & extinct \\
            & 7 & may & allows & fin & mya & deity & called \\
            & 8 & one & air & law & later & ino & used \\
            & 9 & animals & feed & new & months & came & islands \\
            & 10& use & bony & conservation & mysticetes & made & genus \\
            \cmidrule[1pt]{2-8}

            & 0 & dolphins & water & shark & million & poseidon & dolphin \\
            & 1 & various & vertical & corpse & approximately & games & depicted \\
            & 2 & finding & unlike & stocks & assigned & phalanthus & makara \\
            & 3 & military & chew & galea & hybodonts & statue & capensis \\
            & 4 & selfmade & lack & galeomorphii & appeared & isthmian & goddess \\

            \cmidrule{2-8}
            & 0 & species & body & sharks & years & became & meat \\
            & 1 & herd & heart & mostly & acanthodians & pirates & contaminated \\
            & 2 & reproduction & resisting & fda & spent & elder & harpoon \\
            & 3 & afford & fit & lists & stretching & mistook & practitioner \\
            & 4 & maturity & posterior & carcharias & informal & wealthy & pcbs \\

            \bottomrule
    \end{tabular}}
    \caption{For each evaluated matrix factorization method we display the top 10 words for each topic and the 5 most similar words based on cosine similarity for the 2 top words from each topic.}
    \label{table: dolphin_shark_whale_nmf}
\end{table}

\clearpage

\begin{figure}[!ht]
    \begin{minipage}[b]{\linewidth}
    \centering
    \tiny
    \caption*{Articles: ``Soccer'', ``Tennis'', ``Rugby''}
    \resizebox{0.7\textwidth}{!}{%
    \begin{tabular}{lllllll}
        \toprule
        & Topic 1 & Topic 2 & Topic 3 & Topic 4 & Topic 5 & Topic 6 \\
        & \#539 &      \#302 &      \#563 &      \#635 &      \#650 &      \#530 \\
        \midrule
        \multirow{2}{*}{1} & may & leads & tournaments & greatest & football & net \\
        & (0.599) &       (0.212) &      (0.588) &      (0.572) &   (0.553) &   (0.644) \\
        \multirow{2}{*}{2} & penalty & sole & tournament & tennis & rugby & shot \\
        & (0.576) &       (0.205) &      (0.517) &      (0.497) &   (0.542) &   (0.629) \\
        \multirow{2}{*}{3} & referee & competes & events & female & south & stance \\
        & (0.564) &       (0.205) &      (0.509) &       (0.44) &   (0.484) &   (0.553) \\
        \multirow{2}{*}{4} & team & extending & prize & ever & union & stroke \\
        & (0.517) &       (0.204) &      (0.501) &      (0.433) &    (0.47) &   (0.543) \\
        \multirow{2}{*}{5} & goal & fixing & tour & navratilova & wales & serve \\
        & (0.502) &       (0.203) &      (0.497) &      (0.405) &   (0.459) &   (0.537) \\
        \multirow{2}{*}{6} & kick & triggered & money & modern & national & rotation \\
        & (0.459) &       (0.203) &      (0.488) &      (0.401) &   (0.446) &   (0.513) \\
        \multirow{2}{*}{7} & play & bleeding & cup & best & england & backhand \\
        & (0.455) &       (0.202) &      (0.486) &        (0.4) &   (0.438) &   (0.508) \\
        \multirow{2}{*}{8} & ball & fraud & world & wingfield & new & hit \\
        & (0.452) &       (0.202) &      (0.467) &      (0.394) &   (0.416) &   (0.507) \\
        \multirow{2}{*}{9} & offence & inflammation & atp & sports & europe & forehand \\
        & (0.444) &       (0.202) &      (0.464) &       (0.39) &   (0.406) &   (0.499) \\
        \multirow{2}{*}{10} & foul & conditions & men & williams & states & torso \\
        & (0.443) &       (0.201) &      (0.463) &      (0.389) &   (0.404) &   (0.487) \\
        \midrule[1pt]
        \multirow{2}{*}{0} & may & leads & tournaments & greatest & football & net \\
        & (1.0) &            (1.0) &        (1.0) &        (1.0) &        (1.0) &    (1.0) \\
        \multirow{2}{*}{2} & goal & tiredness & events & female & union & shot \\
        & (0.98) &            (1.0) &      (0.992) &       (0.98) &       (0.98) &  (0.994) \\
        \multirow{2}{*}{3} & play & ineffectiveness & tour & ever & rugby & serve \\
        & (0.959) &            (1.0) &      (0.989) &      (0.971) &      (0.979) &  (0.987) \\
        \multirow{2}{*}{4} & penalty & recommences & money & navratilova & association & hit \\
        & (0.954) &            (1.0) &      (0.986) &      (0.967) &       (0.96) &  (0.984) \\
        \multirow{2}{*}{5} & team & mandated & prize & tennis & england & stance \\
        & (0.953) &            (1.0) &      (0.985) &      (0.962) &      (0.958) &  (0.955) \\
        \cmidrule{2-7}
        \multirow{2}{*}{0} & penalty & sole & tournament & tennis & rugby & shot \\
        & (1.0) &            (1.0) &        (1.0) &        (1.0) &        (1.0) &    (1.0) \\
        \multirow{2}{*}{2} & referee & discretion & events & greatest & football & net \\
        & (0.985) &            (1.0) &       (0.98) &      (0.962) &      (0.979) &  (0.994) \\
        \multirow{2}{*}{3} & kick & synonym & event & female & union & serve \\
        & (0.985) &            (1.0) &      (0.978) &      (0.953) &      (0.975) &  (0.987) \\
        \multirow{2}{*}{4} & offence & violated & atp & year & england & hit \\
        & (0.982) &            (1.0) &      (0.974) &      (0.951) &      (0.961) &  (0.983) \\
        \multirow{2}{*}{5} & foul & layout & money & navratilova & wales & stance \\
        & (0.982) &            (1.0) &      (0.966) &      (0.949) &      (0.949) &   (0.98) \\
        \bottomrule
    \end{tabular}}
    \caption{Top half lists the top 10 representative words per dimension of the basis matrix $A$, bottom half lists the 5 most similar words based on cosine similarity for the 2 top words from each topic.}

        \captionof{table}{Top half lists the top 10 representative words per dimension of the basis matrix $A$, bottom half lists the 5 most similar words based on cosine similarity for the 2 top words from each topic.}
        \label{table: soccer_tennis_rugby_dedicom}
    \end{minipage}
    \hspace{1cm}
    \begin{minipage}[b]{\linewidth}
        \centerfloat
          \begin{subfigure}[b]{0.18\textwidth}
             \includegraphics[page=7, width=\linewidth]{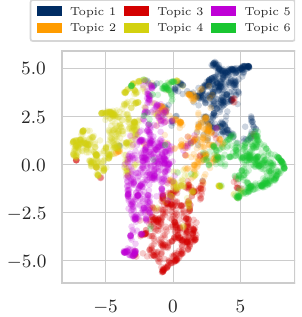}
            \caption{}
            \label{fig:scatter_topics_dedicom_soccer_tennis_rugby}
          \end{subfigure}
          \begin{subfigure}[b]{0.18\textwidth}
            \includegraphics[page=8, width=\linewidth]{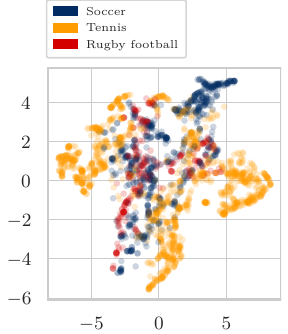}
            \caption{}
            \label{fig:scatter_titles_dedicom_soccer_tennis_rugby}
          \end{subfigure}
          \begin{subfigure}[b]{0.22\textwidth}
            \includegraphics[page=9, width=\linewidth]{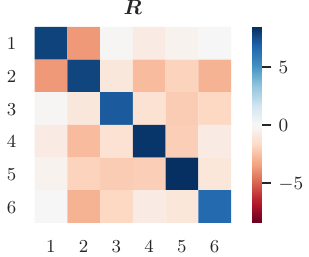}
            \caption{}
            \label{fig:heatmap_dedicom_soccer_tennis_rugby}
          \end{subfigure}
         \caption{(a) 2-dimensional representation of word embeddings $\mat{A}'$ colored by topic assignment. (b) 2-dimensional representation of word embeddings $\mat{A}'$ colored by original Wikipedia article assignment (words that occur in more than one article are excluded). (c) Colored heatmap of affinity matrix $\mat{R}$.}
         \label{fig:plots_soccer_tennis_rugby}
    \end{minipage}
\end{figure}
\clearpage
\begin{table}[!ht]
    \scriptsize
    \centering
    \label{topics_nmf_h}
    \caption*{Articles: ``Soccer'', ``Tennis'', ``Rugby''}
    \resizebox{0.84\textwidth}{!}{%
    \begin{tabular}{llllllll}
        \toprule
        &   & Topic 1 & Topic 2 & Topic 3 & Topic 4 & Topic 5 & Topic 6 \\
        \midrule[1pt]
        & & \#511 &      \#453 &      \#575 &      \#657 &      \#402 &      \#621 \\
        \multirow{20}{*}{NMF} & 1  & net & referee & national & tournaments & rackets & rules \\
        & 2  & shot & penalty & south & doubles & balls & wingfield \\
        & 3  & serve & may & football & singles & made & december \\
        & 4  & hit & kick & cup & events & size & game \\
        & 5  & stance & card & europe & tour & must & sports \\
        & 6  & stroke & listed & fifa & prize & strings & lawn \\
        & 7  & backhand & foul & union & money & standard & modern \\
        & 8  & ball & misconduct & wales & atp & synthetic & greek \\
        & 9  & server & red & africa & men & leather & fa \\
        & 10 & service & offence & new & grand & width & first \\
        \cmidrule[1pt]{2-8}

        & 0  & net & referee & national & tournaments & rackets & rules \\
        & 1  & defensive & retaken & serbia & bruno & pressurisation & collection \\
        & 2  & closer & interference & gold & woodies & become & hourglass \\
        & 3  & somewhere & dismissed & north & eliminated & equivalents & unhappy \\
        & 4  & center & fully & headquarters & soares & size & originated \\
        \cmidrule{2-8}
        & 0  & shot & penalty & south & doubles & balls & wingfield \\
        & 1  & rotated & prior & asian & combining & express & experimenting \\
        & 2  & execute & yellow & argentina & becker & oz & llanelidan \\
        & 3  & strive & duration & la & exclusively & bladder & attended \\
        & 4  & curve & primary & kong & woodbridge & length & antiphanes \\
        \midrule[1pt]

        & &  \#413 &      \#518 &      \#395 &      \#776 &      \#616 &      \#501 \\
        \multirow{20}{*}{LDA} & 1  & used & net & wimbledon & world & penalty & clubs \\
        & 2  & forehand & ball & episkyros & cup & score & rugby \\
        & 3  & use & serve & occurs & tournaments & goal & schools \\
        & 4  & large & shot & grass & football & team & navratilova \\
        & 5  & notable & opponent & roman & fifa & end & forms \\
        & 6  & also & hit & bc & national & players & playing \\
        & 7  & western & lines & occur & international & match & sport \\
        & 8  & twohanded & server & ad & europe & goals & greatest \\
        & 9  & doubles & service & island & tournament & time & union \\
        & 10 & injury & may & believed & states & scored & war \\
        \cmidrule[1pt]{2-8}

        & 0  & used & net & wimbledon & world & penalty & clubs \\
        & 1  & seconds & mistaken & result & british & measure & sees \\
        & 2  & restrictions & diagonal & determined & cancelled & crossed & papua \\
        & 3  & although & hollow & exists & combined & requiring & admittance \\
        & 4  & use & perpendicular & win & wii & teammate & forces \\
        \cmidrule{2-8}
        & 0  & forehand & ball & episkyros & cup & score & rugby \\
        & 1  & twohanded & long & roman & multiple & penalty & union \\
        & 2  & grips & deuce & bc & inline & bar & public \\
        & 3  & facetiously & position & island & fifa & fouled & took \\
        & 4  & woodbridge & allows & believed & manufactured & hour & published \\

        \midrule[1pt]
        & & \#1310 &      \#371 &      \#423 &      \#293 &      \#451 &      \#371 \\
        \multirow{20}{*}{SVD} & 1  & players & net & tournaments & stroke & greatest & balls \\
        & 2  & player & ball & singles & forehand & ever & rackets \\
        & 3  & tennis & shot & doubles & stance & female & size \\
        & 4  & also & serve & tour & power & wingfield & square \\
        & 5  & play & opponent & slam & backhand & williams & made \\
        & 6  & football & may & prize & torso & navratilova & leather \\
        & 7  & team & hit & money & grip & game & weight \\
        & 8  & first & service & grand & rotation & said & standard \\
        & 9  & one & hitting & events & twohanded & serena & width \\
        & 10 & rugby & line & ranking & used & sports & past \\
        \cmidrule[1pt]{2-8}

        & 0  & players & net & tournaments & stroke & greatest & balls \\
        & 1  & breaking & pace & masters & rotates & lived & panels \\
        & 2  & one & reach & lowest & achieve & female & sewn \\
        & 3  & running & underhand & events & face & biggest & entire \\
        & 4  & often & air & tour & adds & potential & leather \\
        \cmidrule{2-8}
        & 0  & player & ball & singles & forehand & ever & rackets \\
        & 1  & utilize & keep & indian & twohanded & autobiography & meanwhile \\
        & 2  & give & hands & doubles & begins & jack & laminated \\
        & 3  & converted & pass & pro & backhand & consistent & wood \\
        & 4  & touch & either & rankings & achieve & gonzales & strings \\
        \bottomrule
    \end{tabular}}
    \caption{For each evaluated matrix factorization method we display the top 10 words for each topic and the 5 most similar words based on cosine similarity for the 2 top words from each topic.}
    \label{table: soccer_tennis_rugby_nmf}
\end{table}

\end{document}